# A visual encoding model based on deep neural networks and transfer learning

Chi Zhang[1], Kai Qiao[1], Linyuan Wang[1], Li Tong[1], Guoen Hu[1], Ruyuan Zhang [2*], Bin Yan[1*]

[1]National Digital Switching System Engineering and Technological Research Center, Zhengzhou, China, 450000

[2]Center for Magnetic Resonance Imaging, Department of Neuroscience, University of Minnesota at Twin Cities, MN. USA. 55108

* co-senior authors

Correspondence:

Ruyuan Zhang

Center for Magnetic Resonance Research

University of Minnesota

2021 6h St SE

Minneapolis, MN 55455-3007

585-752-6673

zhan1217@umn.edu

Bin Yan

National Digital Switching System Engineering and Technological Research Center

No.7, Jianxue Street, Wenhua Road

Zhengzhou City, China, 450000

0086-0371-5598080

ybspace@hotmail.com

**Abstract**

*Background*: Building visual encoding models to accurately predict visual responses is a central challenge for current vision-based brain-machine interface techniques. To achieve high prediction accuracy on neural signals, visual encoding models should include precise visual features and appropriate prediction algorithms. Most existing visual encoding models employ hand-craft visual features (e.g., Gabor wavelets or semantic labels) or data-driven features (e.g., features extracted from deep neural networks (DNN)). They also assume a linear mapping between feature representation to brain activity. However, it remains unknown whether such linear mapping is sufficient for maximizing prediction accuracy.

*New Method*: We construct a new visual encoding framework to predict cortical responses in a benchmark functional magnetic resonance imaging (fMRI) dataset. In this framework, we employ the transfer learning technique to incorporate a pre-trained DNN (i.e., AlexNet) and train a nonlinear mapping from visual features to brain activity. This nonlinear mapping replaces the conventional linear mapping and is supposed to improve prediction accuracy on brain activity.

*Results*: The proposed framework can significantly predict responses of over 20% voxels in early visual areas (i.e., V1-lateral occipital region, LO) and achieve unprecedented prediction accuracy.

*Comparison with Existing Methods*: Comparing to two conventional visual encoding models, we find that the proposed encoding model shows consistent higher prediction accuracy in all early visual areas, especially in relatively anterior visual areas (i.e., V4 and LO).

*Conclusions*: Our work proposes a new framework to utilize pre-trained visual features and train non-linear mappings from visual features to brain activity.

**Highlights:**
- Transfer learning based on DNN can be successfully applied to visual encoding models based on limited fMRI data.
- The dynamic loss function is successfully used for training the ROI-wise visual encoding model.
- Nonlinear mapping can significantly improve the prediction accuracy of the visual encoding model, especially in high-level visual areas.

# 1. Introduction

Vision is the major sensory channel to acquire external information. Characterizing the relationship between input stimulus and endowed human brain activity is not only an important topic in contemporary neuroscience, but also holds the key to promote applications in neural engineering, such as brain-machine interface. Visual encoding models describe how the human brain activity in response to corresponding external stimulus features ([Naselaris T et al. 2011](); [Chen M et al. 2014](); [van Gerven MAJ 2017]()).

The visual processing in the brain is usually in a non-linear fashion. Visual encoding models describe such nonlinear mapping between input visual stimuli and brain activity. So far, visual encoding models based on functional magnetic resonance imaging (fMRI) generally consists of three spaces and two in-between mappings (Fig. 1A) ([Naselaris T et al. 2011](); [Chen M et al. 2014]()). The three spaces are the input stimulus space, the feature space and the brain activity space. The first mapping between the stimulus space and the feature space is usually nonlinear, and this mapping can be regarded as the process of feature extraction. The second mapping between the feature space and the brain activity space is usually linear because this makes it easier to infer the unit tuning along the cortical processing hierarchy in the brain.

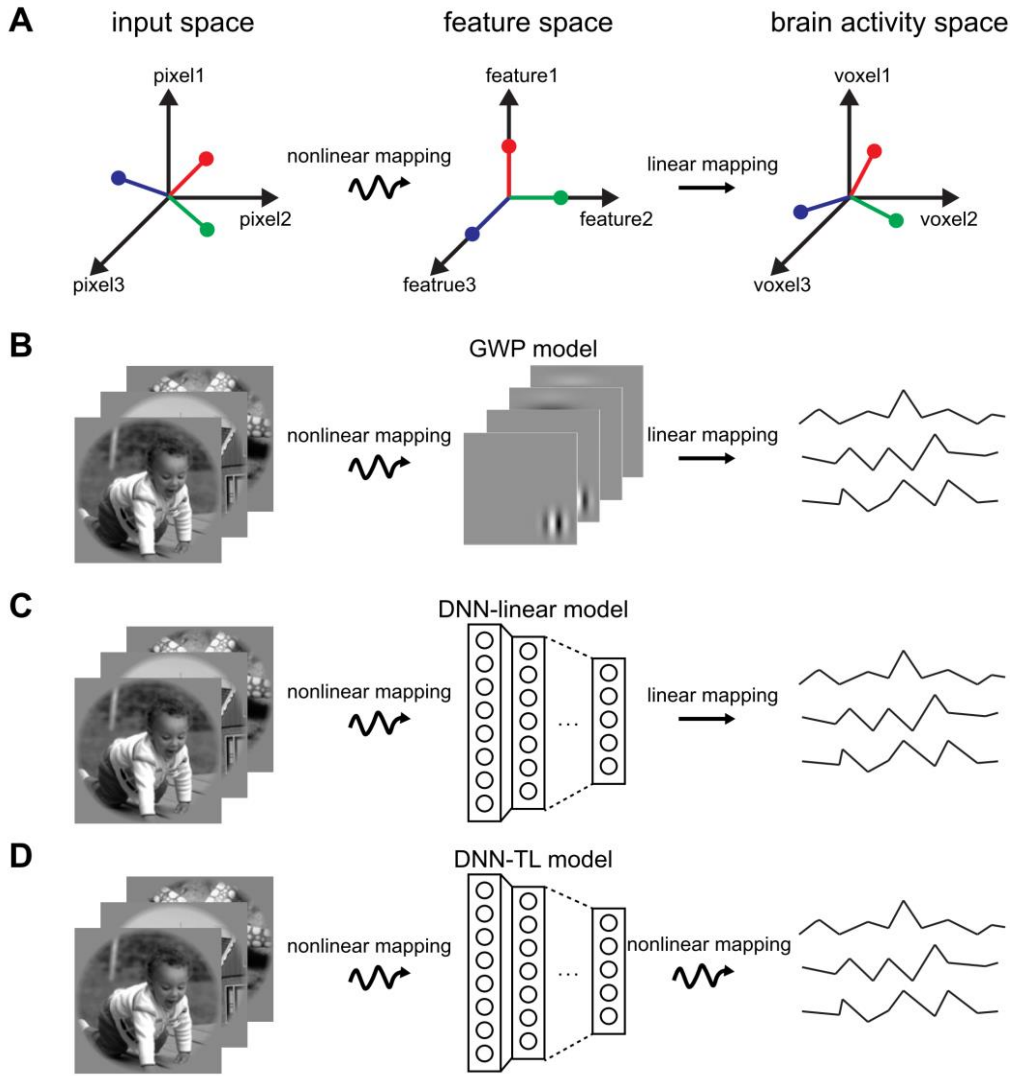

Figure 1. Visual encoding models. *A*. The general architecture of visual encoding models in neuroscience literature. It consists of three spaces (the input space, the feature space and the brain activity space) and two in-between mappings. The mapping from the input space to the feature space is typically nonlinear, which is considered as a process of visual feature extraction. The mapping from the feature space to the brain activity space is typically linear, aimed at linking the brain activity evoked by the visual features. *B.* The architecture of the GWP model. Gabor wavelet pyramid basis functions are used to construct the feature representation in the feature space. The mapping from the feature space to the brain activity space is linear. *C*. The architecture of the DNN-linear model. This model uses the pre-trained features in a DNN (i.e., AlexNet) to construct the feature space. The mapping from the feature space to the brain activity space is linear. *D*. The architecture of the DNN-TL model. The feature space is also contrasted by the pre-trained features in a DNN, but the mapping from the feature space to the brain activity space is nonlinear, implemented by transfer learning.

Earlier visual encoding models are based on some existing computer vision or neuroscience knowledge. In particular, Kay et al. (Kay KN et al. 2008) used the Gabor wavelet pyramid model as the nonlinear feature extractor from the stimulus space to the feature space. Gabor wavelets have been shown as a good approximation of early visual processing (Adelson EH and JR Bergen 1985; Jones JP and LA Palmer 1987; Carandini M et al. 2005). For high-level visual features, visual encoding models are usually based on semantic information (Naselaris T et al. 2009; Huth AG et al. 2012). It depends on the category labels of objects or natural scenes and can predict brain responses in high-level visual areas. Importantly, these studies typically employed hand-crafted features.

Instead of hand-crafted features, deep neural networks (DNN) have been shown to automatically learn hierarchical visual features from a large number of natural images (Krizhevsky A et al. 2012). DNNs have been used to explain both human fMRI data (Yamins DL and JJ DiCarlo 2016; Baker CI and M van Gerven 2018; Tripp B 2018) and monkey neurophysiological data (Yamins DL et al. 2014). With the development of DNN, some new network architectures, such as ResNet (Wen H et al. 2018), recurrent neural network (Shi J et al. 2018), variational autoencoder (Han K et al. 2017) and Capsule Network (Qiao K et al. 2018), have been applied in visual encoding models and have achieved excellent performance.

From the perspective of neural engineering, such as brain-machine interface, accurate brain decoding requires visual encoding models that contain appropriate visual features and prediction algorithms. The existing visual encoding models based on DNN tackling these two problems by using pre-trained layer features and linear mappings (Fig. 1C). The ideal solution is to directly train a deep neural network that takes image stimuli as input and predict the evoked cortical responses. However, it is infeasible since the amount of fMRI data in an empirical experiment is usually scares compared to the large number of parameters in a DNN. Here, we constructed a new visual encoding framework using transfer learning, a deep learning technique aimed at solving the problem of a small amount of data in the target domain (e.g., visual encoding) by using the knowledge acquired in the source domain (e.g., image classification) (Pan SJ and Q Yang 2010). In this framework, we established transfer learning networks for visual encoding based on the pre-trained layers from AlexNet (Krizhevsky A et al. 2012). We denote this as deep neural network transfer learning (DNN-TL) framework.

In addition, we trained two additional layers on top of pre-trained DNN layers to form a nonlinear mapping between the feature space to the brain activity space. As mentioned above, one signature of existing visual encoding models is the linear mapping between the feature space and the brain activity space (Fig. 1B&C). From the perspective of neuroscience, the mapping must be linear in order to reveal functional correspondences between visual features and brain activity. The linear mapping is optimal for theoretical interpretability but might be suboptimal for maximizing prediction accuracy. Here we replaced the linear mapping with a nonlinear mapping (i.e., fully connected layers, Fig. 2D) in order to fine-tune the network and maximize prediction accuracy.

We also compared this DNN-TL framework to two control models that have been frequently used in neuroscience literature. The first one uses Gabor wavelet pyramid (GWP) as visual features ([Kay KN et al. 2008](#)). The second one, denoted as the DNN-linear model here, also uses pre-trained DNN features but assume a linear mapping from the features to brain activity ([Güçlü U and MA van Gerven 2015](#)). We found that our models performed significantly better in predictions of brain activity in a benchmark fMRI dataset. This result indicates that transfer learning technique might be a viable way to achieve higher precision in future research of brain-machine interface.

## 2. Materials and methods

### 2.1 Experimental data

We validated our approach using the fMRI dataset published in Kay KN et al. (2008). The dataset contains brain responses of two subjects (i.e., S1 and S2) acquired using a 4T INOVA MR scanner (Varian, Inc., Palo Alto, CA, USA). The estimated blood oxygen level dependent (BOLD) responses were directly used in training and validating our model, which can be downloaded from an online database (http://crcns.org/data-sets/vc/vim-1). For each subject, there were 1750 stimulus responses for estimation and 120 stimulus responses for validation. The stimuli were grayscale natural images (128 × 128 pixels). The evoked responses in V1, V2, V3, V4 and lateral occipital (LO) area were selected for further analysis. Figures in this study refer to the data from subject 1, consistent results were obtained for subject 2, as shown in Appendix A.

### 2.2 The overview of DNN-TL model

The general transfer learning process is as follows: (1) train a DNN (denoted as base network) in a source domain; (2) copy its first $n$ layers to the first $n$ layers of the target network in a target domain; (3) randomly initialize the remaining layers of the target network; (4) train the target network in the target domain driven by a target task. In step 4, the first $n$ layers of the target network can be frozen or fine-tuned, depending on the size of the target dataset and the number of parameters in the first $n$ layers. We chose the frozen method because of the small amount of fMRI data.

The building process of our DNN-TL modeling in this paper is as follows. First, we used AlexNet as our base network. This well-established base network has been trained on the ImageNet database (Deng J et al. 2009) to classify an image into one of 1000 categories. AlexNet consists of eight layers: five convolutional layers (with 96, 256, 384, 384, and 256 kernels, respectively) and three fully connected layers (4096, 4096, and 1000 artificial neurons). We built eight transfer learning networks for each of the five visual regions of interest (ROI). The eight networks inherited the structure and model parameters from the first one (i.e., layer 1) to the first eight layers (i.e., layer 1-8) of the base network respectively. These inherited layers were then connected to two additional fully connected layers. In these newly established encoding networks, only the parameters of the last two fully connected layers were trainable, ensuring that these networks could be trained on a small number of fMRI data. The whole procedure of the DNN-TL modeling is shown in Fig. 2.

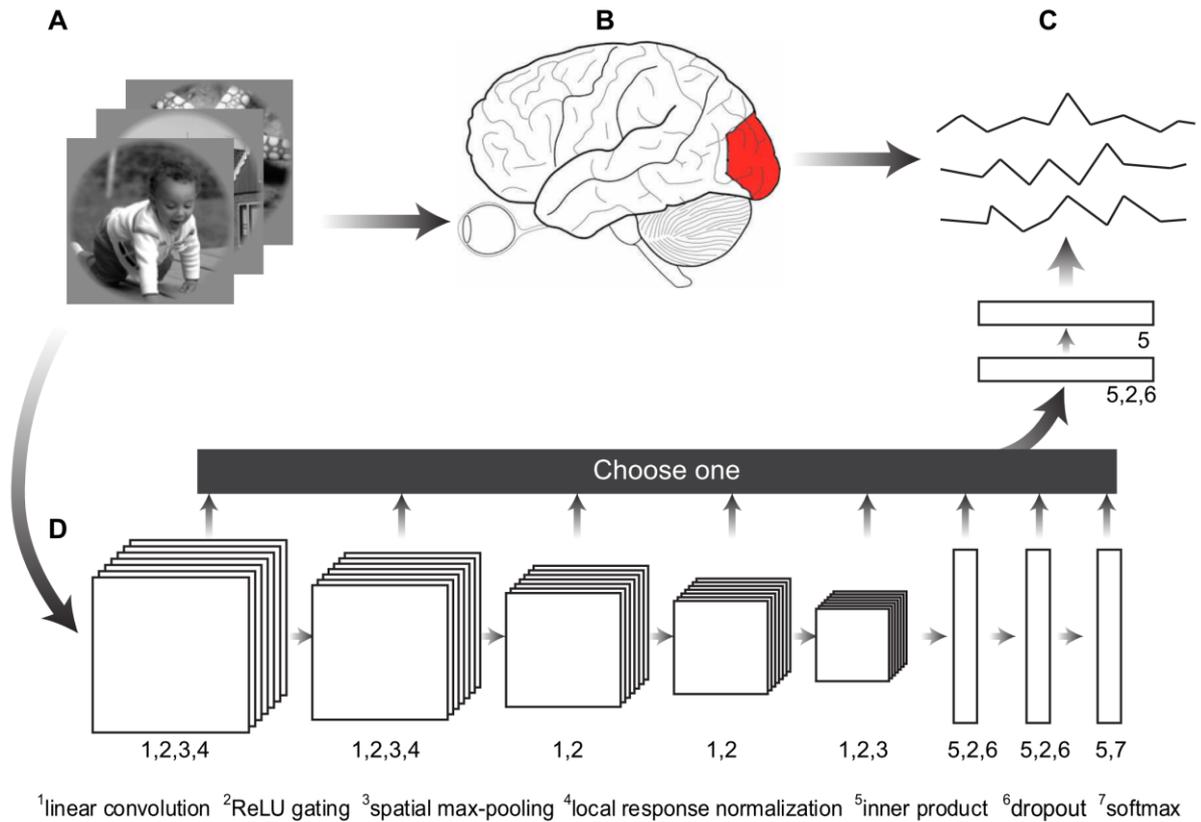

Figure 2. Visual encoding models based on DNN and transfer learning. When a subject is seeing the stimuli (***A***), information flows along the visual pathway (***B***) and evokes visual cortical responses (***C***). A transfer learning framework based on DNN (DNN-TL) is illustrated here to predict visual responses (***D***). Eight transfer learning networks are established by using the first $n$ ($n=1,2\cdots 8$) layers of AlexNet and appending two fully connected layers. The number under each layer indicates the operations used in this layer.

## 2.3 Training DNN-TL models

All existing visual encoding models are voxel-wise encoding models (Naselaris T *et al.* 2011; Chen M *et al.* 2014; van Gerven MAJ 2017). Voxel-wise encoding models can obtain the best model for each voxel based on a training dataset. However, voxel-wise modeling can only use a linear mapping between the feature and the brain activity spaces as it is computationally difficult to train a complex nonlinear model separately for each voxel. In contrast, here we employed an ROI-wise encoding modeling approach—we trained one model to predict the responses of all voxels in each ROI. This approach greatly reduces the computational burden in training, and more importantly, the model parameters are jointly constrained by the prediction accuracy on all voxels. This approach might also render the training process more vulnerable to some noisy voxels. In order to minimize the influences of noisy voxels, we adjusted the loss function dynamically during the training process so as to select voxels that are truly informative. Let us write the trainable parameters in a network concisely as $\Theta = (W, b)$, then the loss function is designed as follows:

$$J(\Theta) = -\sum_v \mu_v C_v(r_m, r_p(\Theta)) + \lambda \|W\|_2^2 \tag{1}$$

where $W$ and $b$ are the weights and bias parameters in the two fully connected layers (Fig. 2). $v$ represents the index of one voxel in a ROI, $C_v(r_m, r_p(\Theta))$ represents the correlation value (Pearson correlation) of empirically measured voxel response $r_m$ and predicted voxel response $r_p(\Theta)$ for the *v*-th voxel. $\mu_v$ represents weighted coefficient for the *v*-th voxel. $\lambda$ is the weighted coefficient of the L2 regularization term. During the training process, we adjusted $\mu_v$ dynamically in order to let the model automatically select those informative voxels. Specifically, the dynamical adjustment was implemented as follows: 1) The training data were randomly divided into the training set containing 1630 images and the validation set containing 120 images. It should be noted that the validation set was different from the test set (i.e., 1630 images in the training set, 120 images in the validation set and 120 images in the test set), which was used to calculate the prediction accuracy (see section **2.5**); 2) Initialization. Let $\mu_v$ be the constant 1, and first trained one epoch on the training set, and got $\Theta$; 3) Calculated $r_p(\Theta)$ on the validation set and got $C_v$, then $\mu_v$ was updated by the following function:

$$\mu_v = \begin{cases} 0 & (C_v < 0) \\ C_v / 0.27 & (0 \leq C_v < 0.27) \\ 1 & (C_v \geq 0.27) \end{cases} \tag{2}$$

4) Fixed $\mu_v$ and trained an epoch on the training set to get new $W$ and $b$; 5) Repeated steps 3) 4) until the loss converged.

In this way, for each ROI, we build eight encoding models from layer 1 to layer 8. For each voxel, the model with the highest encoding accuracy was chosen as the final model. The value of 0.27 in the function when updating $\mu_v$ was obtained by randomization test at $p < 0.001$ (see below). The model was built and solved using TensorFlow ([Abadi M et al. 2016](#)).

## 2.4 Control models

In order to further validate the performance of our model, we compared it with two control encoding models. The first model is the classic GWP model (Fig. 1B) ([Kay KN et al. 2008](#)).

The GWP model is a voxel-wise encoding model that uses Gabor wavelet pyramid basis functions construct the feature representations in the feature space. Concretely, we used 48 Gabors at 6 log-spaced spatial frequencies from 1 cycle/field-of-view (FOV) to 32 cycles/FOV. For each frequency, we sampled 8 evenly spaced orientations between 0 and $7\pi/8$. The responses of the GWP model were defined as the square root of the concentrated energy of the quadrature phase wavelets with the same position, direction and spatial frequency. The mapping from the feature space to the brain activity space was implemented using sparse linear regression. We used regularized orthogonal matching pursuit (ROMP) (Needell D and R Vershynin 2009; 2010) as the estimation method, which is the same as our previous research (Zhang C et al. 2018).

The second encoding model is based on features in a DNN (DNN-linear) (Fig. 1C), which is similar to the method described in Güçlü U and MA van Gerven (2015). This model uses the pre-trained features in AlexNet as nonlinear feature extractors to construct the feature space. It needs to be emphasized again that the DNN-linear model is also voxel-wise. For each voxel in visual cortex, a linear mapping from the feature space to the brain activity space is trained. This linear mapping uses the same sparse linear regression optimization approach as above. Since AlexNet has 8 layers and in theory any layer can be used as the feature space to predict response, we established 8 encoding models from layer 1 to layer 8 respectively and select the model with the highest prediction accuracy as the final DNN-TL model.

## 2.5 Quantification of model performance

We define the prediction accuracy for a voxel as Pearson correlation between the observed and the predicted responses across all 120 images in the test set. For each voxel, we calculated the prediction accuracy of each of the three models (DNN-TL, GWP and DNN-linear). To compare models, we first made a scatter plot in which each dot corresponds to a single voxel (Fig. 3A&4A). The ordinate value of each dot represents the prediction accuracy of one control model (GWP or DNN-linear), while the abscissa value represents the prediction accuracy of the DNN-TL model. Second, we plotted the distribution of prediction accuracy difference of the voxels on whom both models yielded significant prediction. Here, the correlation threshold for significance prediction is 0.27 ($p < 0.001$, randomization test, see below) (Fig. 3B&4B). Lastly, the voxels in each ROI were sorted in the descending order of the prediction accuracy values (Fig. 5), so as to analyze the relationship between the prediction accuracy and the number of effective encoding voxels in the three models.

To examine whether each voxel's prediction accuracy value significantly deviated from the null hypotheses, we randomly shuffled the pairing between observed and predicted responses across 120 test images in the test set 1000 times and in each randomized sample recalculated the voxel's prediction. This calculation produced a null hypothesis distribution for each voxel. For all voxels, the prediction accuracy value above 0.27 was significant ($p<0.001$). To examine the significance of a model advantage (percent of voxels with higher prediction accuracy), we randomly permuted (with 0.5 probability) the model prediction accuracy of each voxel whose prediction accuracy is above 0.27 for at least one of the two models and recalculated the advantage for each model. We repeated the calculation for 1000 times and obtained a null distribution of model advantage. For any two models, a model advantage deviating from 50% greater than 3% was significant ($p<0.05$) from its null hypothesis distribution.

## 3. Results

### 3.1 Comparisons between the DNN-TL model and the GWP model in prediction accuracy

The prediction accuracy of the DNN-TL model was compared with that of the GWP model (Fig. 3). We found that the DNN-TL model was significantly better than the GWP model in all ROIs. From V1 to LO, the advantages of our model became more and more obvious, from 73.2% to 98.6%. Especially in LO, there were only a few voxels whose responses can be explained significantly by the GWP model and the DNN-TL model almost performed better in all voxels.

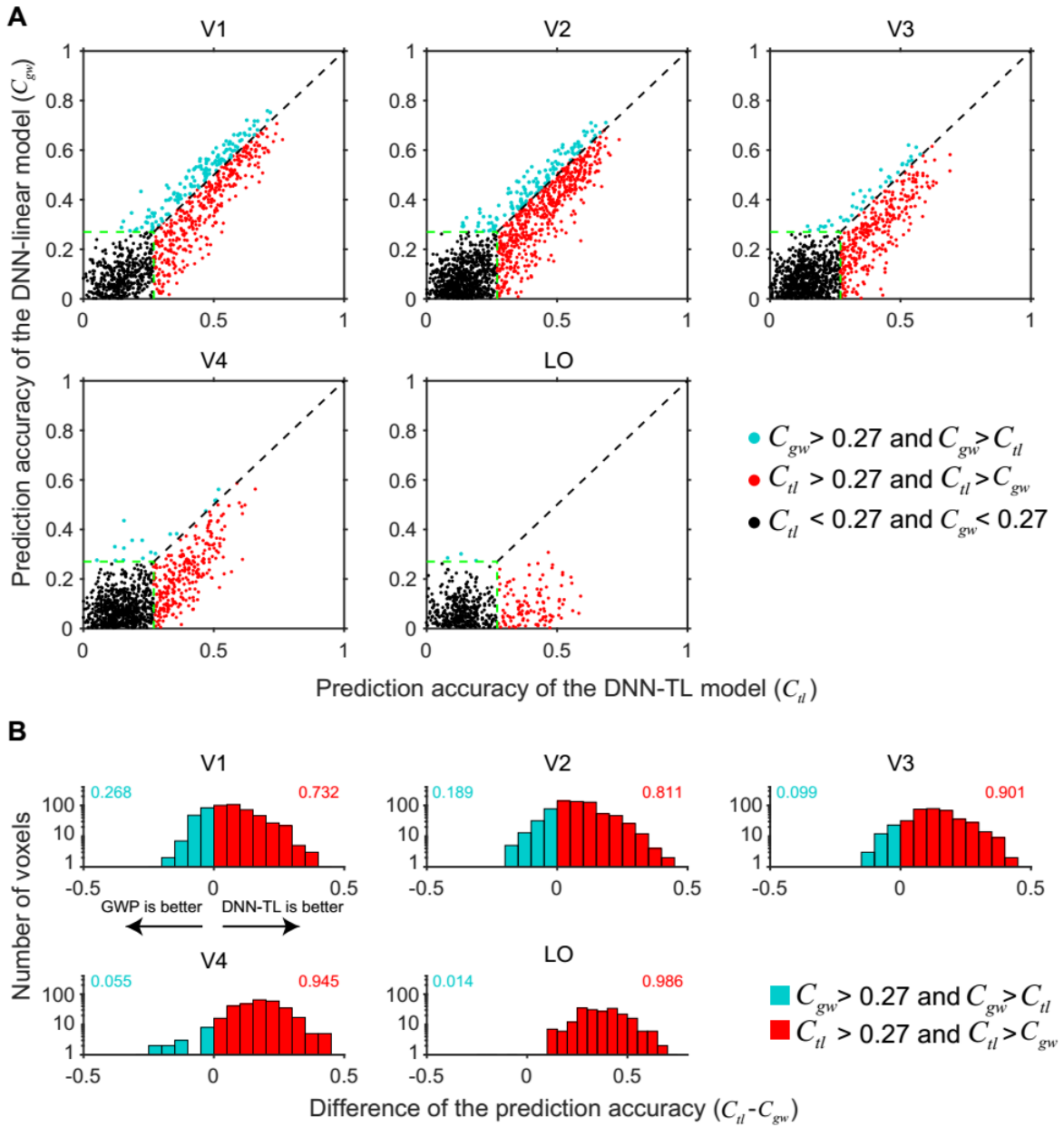

Figure 3. Comparison of prediction accuracy between the DNN-TL ($C_{tl}$) and GWP ($C_{gw}$) models. A. Each of the five axes displays a comparison between the prediction accuracy of the two models in one ROI. In all five scatter plots, the ordinate and abscissa represent the prediction accuracy values of the GWP

model and the DNN-TL model respectively. The green dashed lines indicate the significant prediction value 0.27 (p < 0.001, randomization test). The black dots indicate the voxels on whom the prediction accuracy values of both models are not significant (i.e., under 0.27). The red dots indicate the voxels whose responses can be better predicted by the DNN-TL model than the GWP model and vice versa for the cyan dots. B. Distribution of the prediction accuracy difference between the DNN-TL and GWP models. Prediction accuracy difference above 0 indicates higher prediction accuracy of the DNN-TL model, as marked by red color, and vice versa for the cyan color of the GWP model. The number on each side represents the fraction of voxels whose prediction accuracy are higher under that model.

**3.2 Comparative analysis of the DNN-TL and DNN-linear models in prediction accuracy**

Fig. 4 shows the comparisons between the prediction accuracy values of the DNN-TL and the DNN-linear models. The DNN-TL model has significant advantages over the DNN-linear model in V3, V4 and LO ($p < 0.05$, permutation test), while there was no significant difference between the two models in V1 and V2 ($p > 0.05$, permutation test).

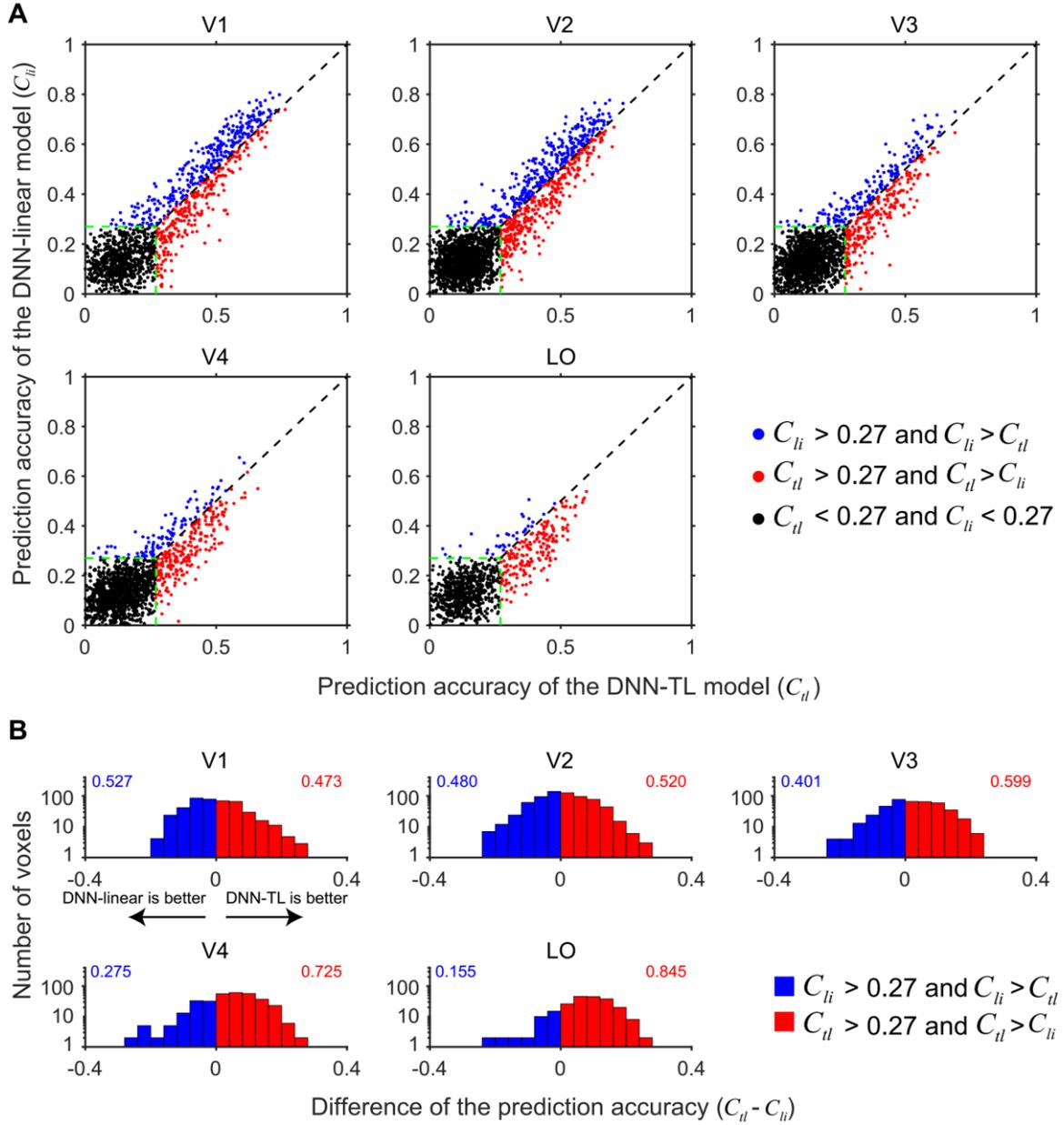

Figure 4. Comparison of prediction accuracy between the DNN-TL ($C_{tl}$) and the DNN-linear ($C_{li}$) models. All symbol definitions in the subplots are the same as Fig. 3, except that here the control model is the DNN-linear model (as marked by blue color).

### 3.3 Model comparison by sorting voxels in prediction accuracy

We extracted the voxels whose responses can be significantly predicted by each of the three models ($C > 0.27$) and sorted them in a descending order of the prediction accuracy (Fig. 5). Overall, the DNN-TL (red lines in Fig. 5) and the DNN-linear (blue lines in Fig. 5) models are better than the GWP model (cyan lines in Fig. 5). Here we focus on comparing the DNN-TL model and the DNN-linear model. The DNN-TL model could significantly predict responses of 39.5%, 31.5%, 21.0%, 19.4% and 22.3% voxels in V1, V2, V3, V4 and LO, respectively. However, in V1, V2 and V3, the prediction accuracy of the top voxels calculated by DNN-linear model is higher than that of DNN-TL. We speculate that there is a strong linear correlation between a small number of voxel responses and DNN features in the low-level visual area and that the use of non-linear mapping will lead to over-fitting and reduce the prediction accuracy.

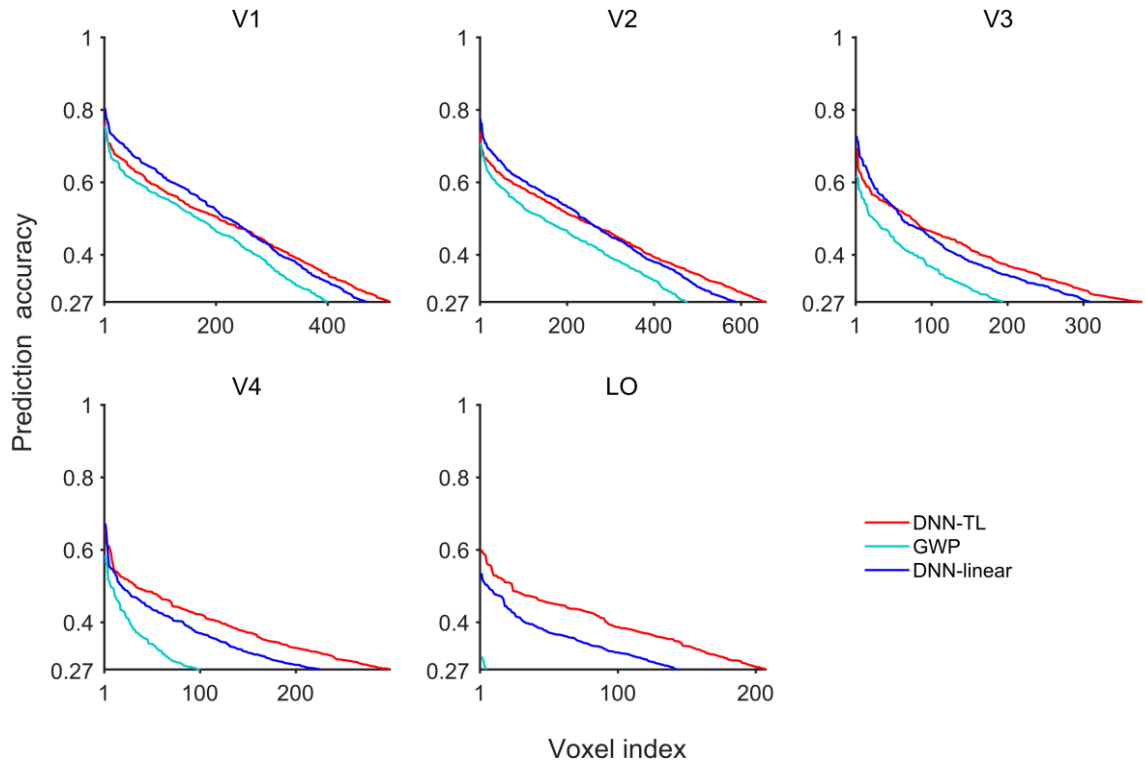

Figure 5. Comparisons of the DNN-TL, GWP, and DNN-linear models by sorting voxels in the descending order of prediction accuracy. Only the voxels on whom the prediction accuracy value exceeds 0.27 are plotted. Red, cyan and blue curves represent the DNN-TL, the GWP and the DNN-linear models, respectively. The DNN-linear model exhibits significantly higher performance in higher level areas, such as V4 and LO.

## 4. Discussion

### 4.1 Features used for visual encoding: accuracy vs. interpretability

The interpretability and accuracy are two important indicators for evaluating an encoding model ([Kay KN 2018](#)). The accuracy indicates to what extent a visual encoding model can predict the brain activity evoked by novel stimuli. The interpretability indicates how well we can capture the relationship between the components of the encoding model (i.e., visual features) and the outcomes that the model predicts (i.e., brain activity).

In previous voxel-wise encoding models, feature extraction is the key to achieve high prediction accuracy ([Naselaris T et al. 2011](#); [van Gerven MAJ 2017](#)). Gabor wavelets are most frequently used features for low-level visual processing but inappropriate for higher level visual processing. The success of deep neural networks in computer vision inspired researchers to utilize features in DNNs since DNNs have been shown to contain the most comprehensive and effective features ranging from low-level to high-level visual processing. These features are believed to be better descriptions of the human brain visual information processing than other hand-craft features. In this paper, the two models using DNN features, the DNN-TL and the DNN-linear models, have higher prediction accuracy than the GWP model in all brain ROIs (V1-LO). This again suggests that DNN can indeed be used as feature extractors in the encoding model.

To ensure high interpretability from the perspective of neuroscience, previous visual encoding models assumed a linear mapping between the feature space to the brain activity space since "linearly decodable" is the key signature that can link visual features to unit tuning ([Hong H et al. 2016](#)). However, the linear mapping might not achieve the best prediction accuracy for engineering purposes. Our results demonstrate that non-linear mapping can give a more accurate prediction.

### 4.2 Mapping from the feature space to the brain activity space: linear vs. non-linear

This paper focuses on the linear and non-linear problems when mapping the feature space to the brain activity space, which is also the essential difference between the DNN-TL and the DNN-linear models. The DNN-TL model can more accurately predict voxel responses than the DNN-linear model. This advantage is more prominent in relatively higher visual areas (i.e., LO). We speculate that the ways that DNNs and the human visual pathway extract visual features diverged more and more from low-level to high-level visual areas. This may be because DNNs are always trained on a single target task (i.e., image classification) and its high-level features are only optimized for this target task. In contrast, visual information in the human brain is usually used to guide multiple tasks (e.g., classification, recognition, detection, location, etc.) at the same time. Therefore, the response in high-level visual areas might contain more complex features than for a single task.

**4.3 Training encoding model: ROI-wise vs. voxel-wise**

In this paper, it is repeatedly emphasized that the DNN-TL model is ROI-wise and the two control models are voxel-wise. The key difference is that the parameters in the DNN-TL model are constrained by all voxels while voxel-wise encoding models are trained separately for each voxel. But it is possible that the model performance of the DNN-TL modeling is drastically perturbed by a small set of noisy or non-informative voxels. To circumvent this, we adaptively adjust the contribution of different voxels towards loss function during the training process. This approach ensures to select the most informative voxels. Our pilot model training shows that the ROI-wise encoding model is less effective if the loss function is not dynamically adjusted. The DNN-TL model can perform much better if it can be trained on individual voxels with sufficient computational power. Such adaptive model training method can be used in future visual modeling practice.

**4.4 Future directions for visual encoding models**

In our training process, we froze the layers adopted from the pre-trained DNN network and only optimized two fully connected layers. This is because of the limited amount of fMRI data and is not the optimal method in theory. If the parameters of the new transfer learning network are fine-tuned on sufficient training data, the performance of the DNN-TL model might be further improved (Yosinski J et al. 2014).

In our previous discussion on linear and non-linear mapping, we pointed out that DNNs differ from human vision in terms of target tasks. Nowadays, almost all encoding models based on DNNs that have been trained on a single task (e.g. classification). Considering the features of DNNs that are trained for other tasks, such as semantic segmentation or target location, it is possible to improve model prediction accuracy by aggregating the features from the models trained for different tasks, since this is more similar to what human brain actually did (Yang GR et al. 2019).

## 5. Conclusion

Building accurate visual encoding models is critical for successful brain decoding. Through the technique of transfer learning, we constructed and trained an encoding model to predict the response in visual cortex. By comparing the prediction accuracy of each visual ROI, we showed that the proposed model can better predict the voxel response in LO compared to early visual areas such as V1 and V2. Future studies need to further explore the feasibility of using pre-trained visual features and training non-linear prediction algorithms on these features.

**Acknowledgments**

This work was supported by the National Key Research and Development Plan of China under grant 2017YFB1002502, the National Natural Science Foundation of China (No.61701089), and the Natural Science Foundation of Henan Province of China (No.162300410333).

**Author contributions**

C.Z., L.W., L.T., G.H., R.Z. and B.Y. developed research idea and study concepts; C.Z., K.Q. performed the data analyses; C.Z., R.Z. wrote the manuscript.

**Competing interests**

The authors declare that they have no competing interests.

## Appendices A. Results for subject 2

The results of subjects 2 were consistent with those of subjects 1. Fig. A1, A2 and A3 correspond to the sections **3.1**, **3.1** and **3.2** in the main text.

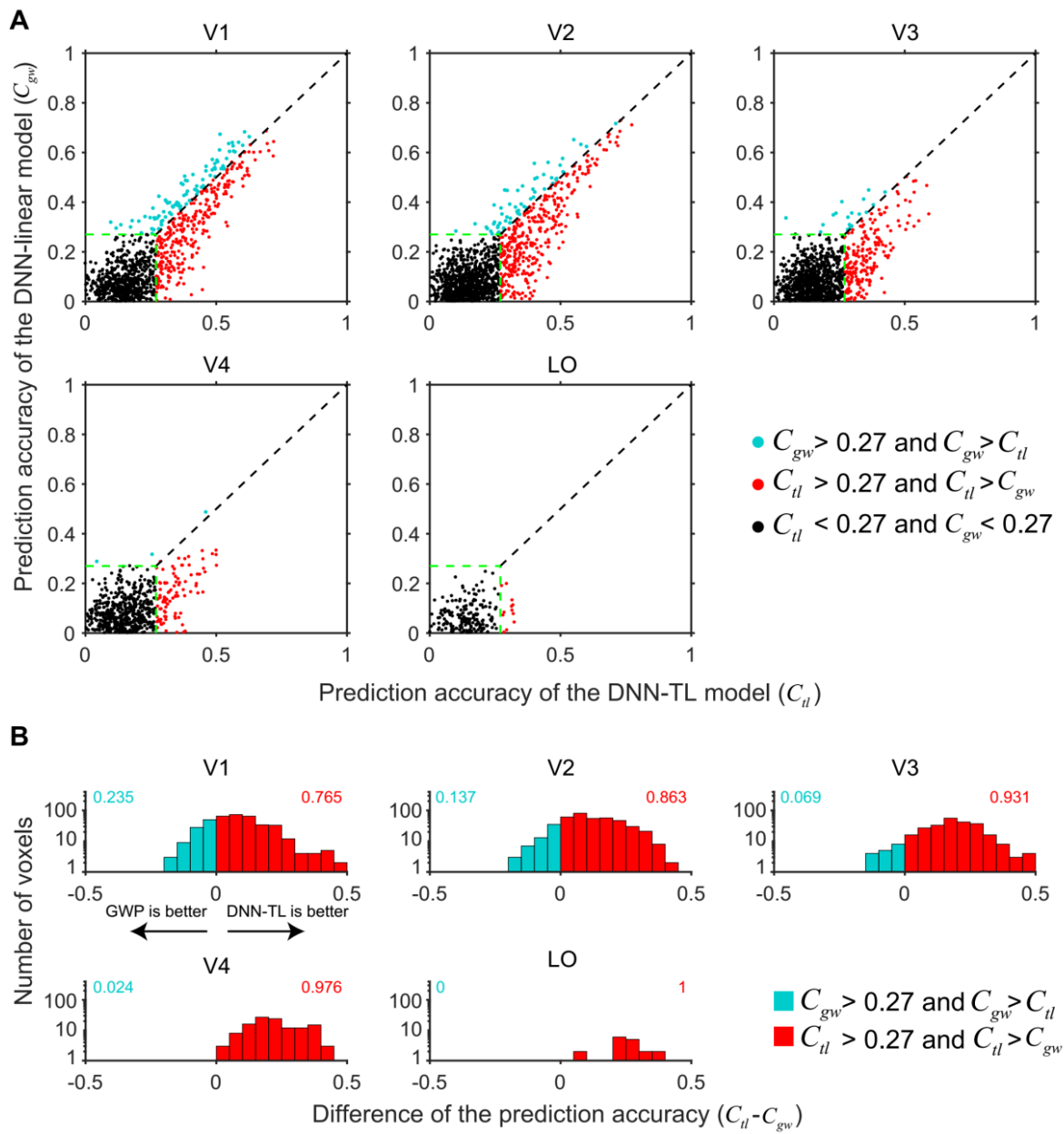

Figure A.1. Comparison of prediction accuracy between the DNN-TL ($C_{tl}$) and GWP ($C_{gw}$) models for subject 2. Refer to Figure 3 for a detailed description of the plot elements.

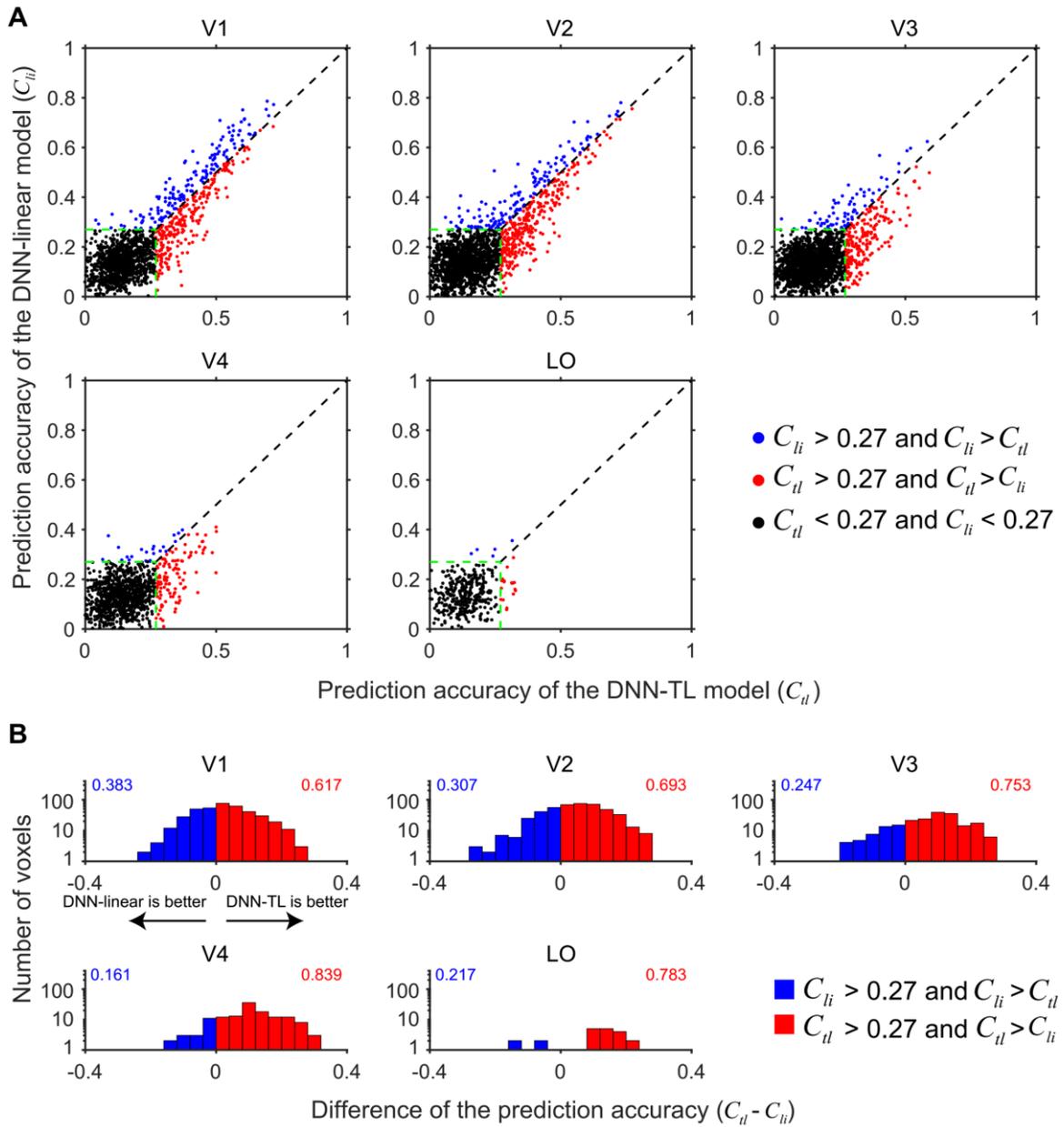

Figure A.2. Comparison of prediction accuracy between the DNN-TL ($C_{tl}$) and the DNN-linear ($C_{li}$) models for subject 2. Refer to Figure 3 for a detailed description of the plot elements.

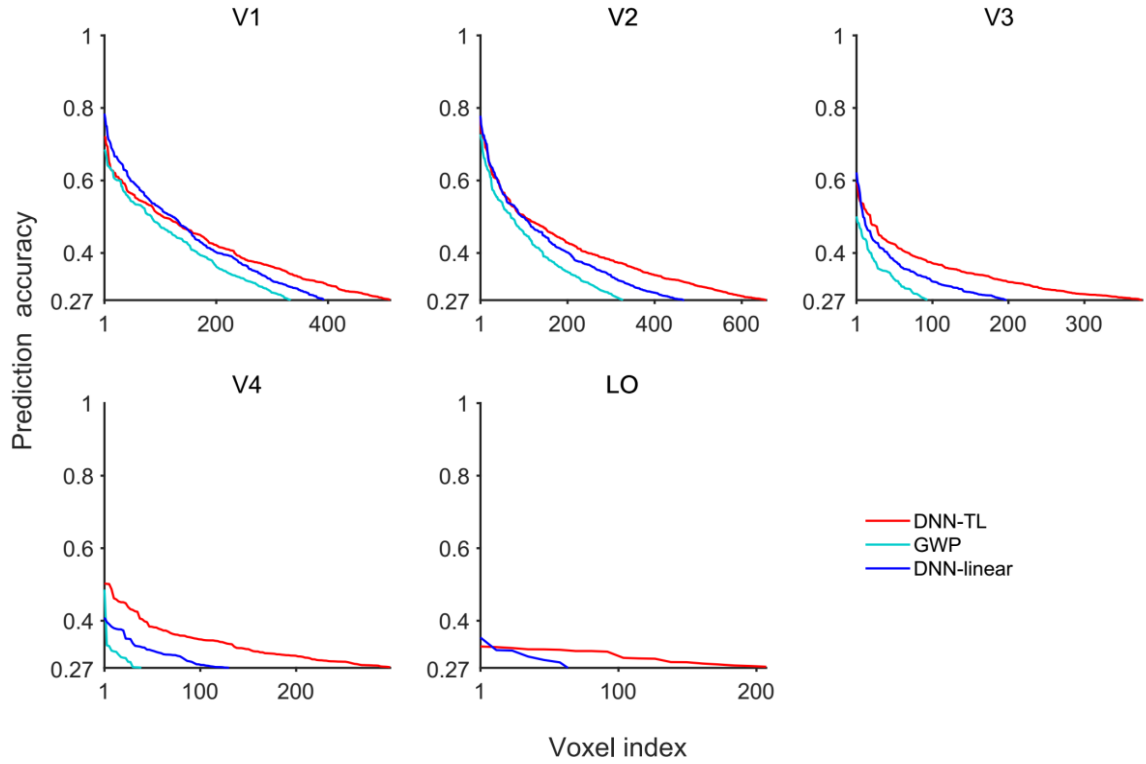

Figure A.3. Comparisons of the DNN-TL, GWP, and DNN-linear models by sorting voxels in the descending order of prediction accuracy for subject 2. Refer to Figure 5 for a detailed description of the plot elements.